\documentclass[sigconf]{acmart}

\usepackage{multirow}
\usepackage{graphicx}
\usepackage{url}
\usepackage{booktabs}
\usepackage{enumitem}
\usepackage{CJKutf8}
\usepackage{natbib}

\usepackage{color,xcolor}
\usepackage{enumitem}

\usepackage{colortbl}
\usepackage{graphicx}
\usepackage{soul}
\usepackage{url}
\usepackage{amsmath}
\usepackage[utf8]{inputenc}
\usepackage{xcolor}
\usepackage{algorithm}

\usepackage{amssymb}
\usepackage{makecell}
\usepackage{xspace}
\usepackage{array}
\usepackage[normalem]{ulem}
\usepackage{footmisc}
\usepackage{colortbl}
\usepackage{epstopdf}
\usepackage{setspace}
\usepackage{pifont}
\usepackage{balance}

\newcommand{\ie}{\emph{i.e.,}\xspace}
\newcommand{\cbkgrnd}{\cellcolor{blue!15}}

\newcommand{\dubbelop}{$^{\blacktriangle}$}

\newcommand{\dubbelneer}{$^{\blacktriangledown}$}

\newcommand{\hlc}[2][yellow]{{%
    \colorlet{foo}{#1}%
    \sethlcolor{foo}\hl{#2}}%
}

\AtBeginDocument{%
  \providecommand\BibTeX{{%
    \normalfont B\kern-0.5em{\scshape i\kern-0.25em b}\kern-0.8em\TeX}}}

\copyrightyear{2023}
\acmYear{2023}
\setcopyright{acmlicensed}\acmConference[WSDM '23]{Proceedings of the Sixteenth ACM International Conference on Web Search and Data Mining}{February 27-March 3, 2023}{Singapore, Singapore}
\acmBooktitle{Proceedings of the Sixteenth ACM International Conference on Web Search and Data Mining (WSDM '23), February 27-March 3, 2023, Singapore, Singapore}
\acmPrice{15.00}
\acmDOI{10.1145/3539597.3570476}
\acmISBN{978-1-4503-9407-9/23/02}

\makeatother

\begin{document}

\begin{CJK*}{UTF8}{gkai}

%\title{Improve Retrieval-Based Chatbots via Dialogue Simulation: \\A General Framework for Multi-Turn Response Selection}
\title{EZInterviewer: To Improve Job Interview Performance with Mock Interview Generator}

\author{Mingzhe Li$^{*\dagger}$}
\affiliation{%
  \institution{Peking University}
%   \country{dd}
}
\email{li_mingzhe@pku.edu.cn}

\author{Xiuying Chen*}
\affiliation{%
  \institution{CBRC, KAUST\\
  CEMSE, KAUST}
}
\email{xiuying.chen@kaust.edu.sa}

\author{Weiheng Liao}
\affiliation{%
  \institution{Made by DATA}
%   \country{dd}
}
\email{Liaoweiheng@gmail.com}

\author{Yang Song}
\affiliation{%
  \institution{BOSS Zhipin NLP Center}
%   \country{dd}
}
\email{songyang@kanzhun.com}

\author{Tao Zhang}
\affiliation{%
  \institution{BOSS Zhipin}
%   \country{dd}
}
\email{kylen.zhang@kanzhun.com}

\author{Dongyan Zhao}
\affiliation{%
  \institution{Peking University}
%   \country{dd}
}
\email{zhaody@pku.edu.cn}

\author{Rui Yan$^{\ddagger}$}
\affiliation{%
  \institution{Gaoling School of AI\\Renmin University of China}
%   \country{dd}
}
\email{ruiyan@ruc.edu.cn}

\def\authors{Mingzhe Li, Xiuying Chen, Weiheng Liao, Yang Song, Tao Zhang, Dongyan Zhao, Rui Yan}

\renewcommand{\shortauthors}{Mingzhe Li et al.}

\thanks{* Both authors contributed equally to this research.\\
$\dagger$ Work done during an internship at BOSS Zhipin.\\
$\ddagger$ Corresponding author: Rui Yan (ruiyan@ruc.edu.cn).} 
%%
%% The abstract is a short summary of the work to be presented in the
%% article.
\begin{abstract}
Interview has been regarded as one of the most crucial step for recruitment.
To fully prepare for the interview with the recruiters, job seekers usually practice with mock interviews between each other.
However, such a mock interview with peers is generally far away from the real interview experience: the mock interviewers are not guaranteed to be professional and are not likely to behave like a real interviewer.
Due to the rapid growth of online recruitment in recent years, recruiters tend to have online interviews, which makes it possible to collect real interview data from real interviewers.
In this paper, we propose a novel application named EZInterviewer, which aims to learn from the online interview data and provides mock interview services to the job seekers.
The task is challenging in two ways: (1) the interview data are now available but still of low-resource; (2) to generate meaningful and relevant interview dialogs requires thorough understanding of both resumes and job descriptions.
To address the low-resource challenge, EZInterviewer is trained on a very small set of interview dialogs.
The key idea is to reduce the number of parameters that rely on interview dialogs by disentangling the knowledge selector and dialog generator so that most parameters can be trained with ungrounded dialogs as well as the resume data that are not low-resource.
Specifically, to keep the dialog on track for professional interviews, we pre-train a knowledge selector module to extract information from resume in the job-resume matching.
A dialog generator is also pre-trained with ungrounded dialogs, learning to generate fluent responses.
Then, a decoding manager is finetuned to combine information from the two pre-trained modules to generate the interview question.
Evaluation results on a real-world job interview dialog dataset indicate that we achieve promising results to generate mock interviews.
With the help of EZInterviewer, we hope to make mock interview practice become easier for job seekers.

%to achieve the same level of performance as state-of-the-art baselines, our model only needs 1/10 training data (∼1.3k interview dialogs).
%And when the full set of training data is used, our model outperforms those baselines by 5.3\% in terms of BLEU-1 metric.

%Online interview services, which are rapidly changing the landscape of the job market, face one major challenge: to conduct effective job interviews at scale.
%While many regard Robotic Mock Interviewer (RMI) as a potential powerful solution to keep human intervention to a minimum, its use is limited by the lack of sufficient training data: resume-grounded interview dialogs are extremely costly to collect.
%In this paper, we propose a \textit{Mock Interview Generation} (MIG) task, the goal of which is to generate relevant, meaningful multi-turn interview dialog with a job candidate based on both of the candidate's resume and the job requirements.

\end{abstract}

%%
%% Keywords. The author(s) should pick words that accurately describe
%% the work being presented. Separate the keywords with commas.
\keywords{EZInterviewer, mock interview generation, knowledge-grounded dialogs, online recruitment, low-resource deep learning}

\begin{CCSXML}
<ccs2012>
   <concept>
       <concept_id>10010147.10010178.10010179.10010182</concept_id>
       <concept_desc>Computing methodologies~Natural language generation</concept_desc>
       <concept_significance>500</concept_significance>
       </concept>
 </ccs2012>
\end{CCSXML}

\ccsdesc[500]{Computing methodologies~Natural language generation}

%%
%% This command processes the author and affiliation and title
%% information and builds the first part of the formatted document.
\maketitle

\section{Introduction}
\begin{figure}
    \centering
    \includegraphics[scale=0.14]{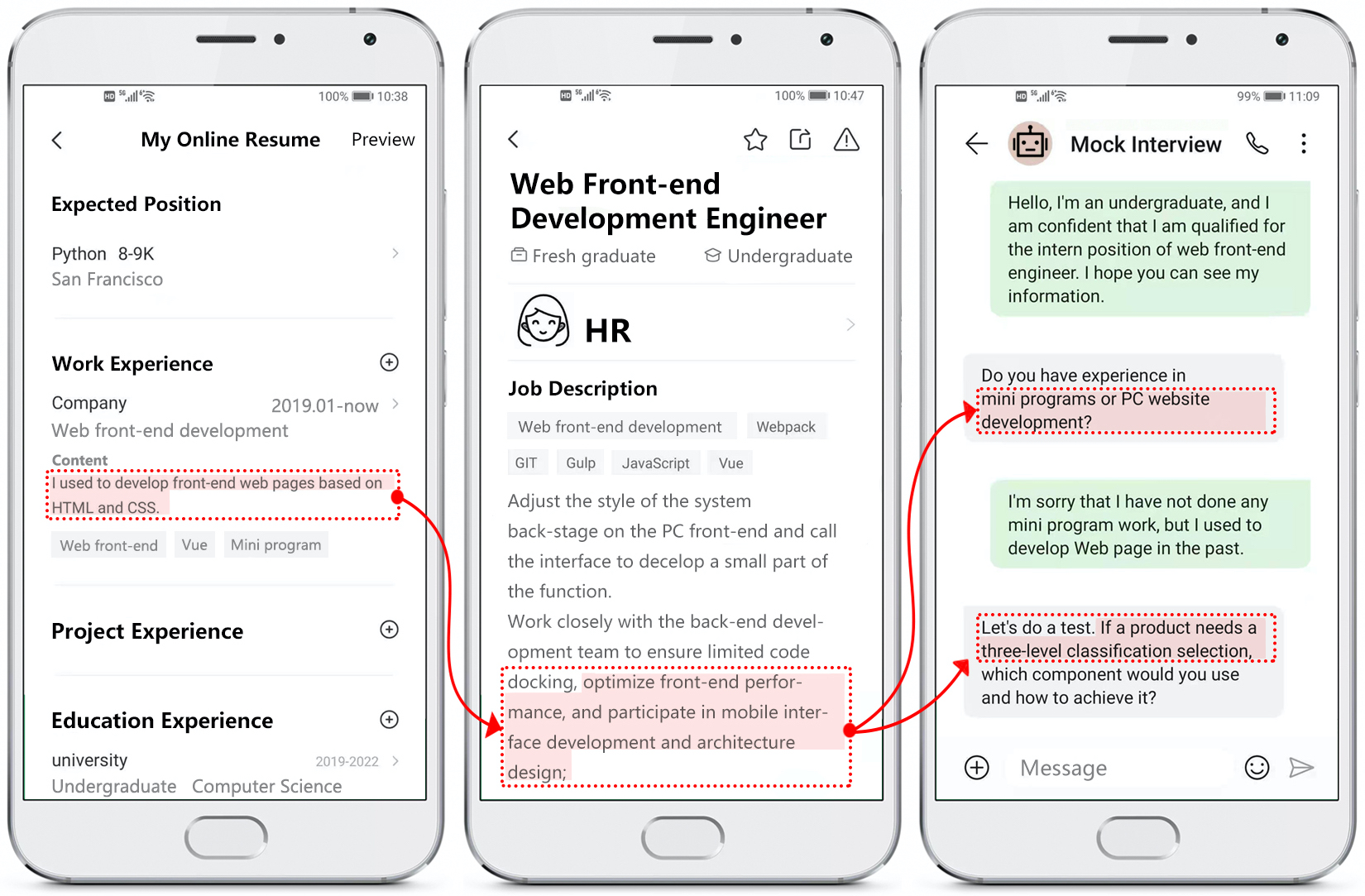}
    \caption{
        An example of the Mock Interview Generation task. Based on the candidate's work experience and the current dialog on the experience of web page development, the system generates an interview question ``If a product needs a three-level classification selection, which component would you use and how to achieve it?''.
    }
    \label{fig:intro}
\end{figure}

% The advent and rapid growth of online recruitment services are transforming the way of how people are being hired in the job market: every day millions of job seekers post their resumes online and search for their next ideal jobs, while as many recruiters are actively on the hunt for suitable candidates.
% When a resume-job description (JD) match happens, a series of interviews would follow, the first round of which is usually conducted online.
% The continuous trend of moving hiring events from off-line to on-line---in particular, accelerated by the current COVID-19 pandemic---means online interviews now become the ``new norm'' for everyone.
% As things are happening online, there is no surprise that job seekers now rely heavily on online sources to prepare for their interviews today. 
% Various websites provide forums for them to share their experience of job interviews, such as GlassDoor~\footnote{\url{https://www.glassdoor.com}} and Boss Zhipin~\footnote{\url{https://www.kanzhun.com}}.
To make better preparations, job seekers practice mock interviews, which aims to anticipate interview questions and prepare them for what they might get asked in their real turn.
% Nevertheless, due to the time and costs associated with hiring professional career counselors, they usually have a friend play the recruiter's part.
However, the outcome of such an approach is unsatisfactory, since those ``mock interviewers'' do not have interview experience themselves, and do not know what the real recruiters would be interested in.
%  /and cannot provide professional feedback to the job seekers within the interview dialogs.
Mock Interview Generation (MIG) represents a plausible solution to this problem. 
Not only makes interviews more cost-effective, but mock interview generators also appear to be feasible, since much can be learned about the job seekers from their resumes, as can the job itself from the job description (JD).
% Hence, a meaningful, professional interview is possible if we can train a mock interview generator to acquire conversational skills and to gain a good understanding of the job and the candidate.
% We may even tell the points of attention as well as the interview style of a recruiter based on his/her online interactions with past candidates.
% Hence a personalized, meaningful interview is possible with an MIG which has a good understanding of the actual interviewer, the job, and the candidate.
% In order to develop an RMI capable of conducting meaningful and effective job interviews, we propose a Mock Interview Generation (MIG) task, the goal of which is of two folds:
% Clearly, the goal of a Mock Interview Generation task is of two folds: to produce professional, relevant interview questions based on the interview dialog contexts to examine the job candidate's abilities given both the candidate's resume and the job requirements; and to engage in a flowing, coherent interview conversation with the job candidate, just as the candidate would expect from a human recruiter.
An illustration of MIG task is shown in Figure~\ref{fig:intro}.
% Based on the work experience in the resume and multi-turn dialog context, an interview happens between job candidate and recruiter.

%[LWH: moved to the later chapters to reduce the length of introduction] Based on the skill information ``Web front-end'', work experience information ``used to develop front-end web pages based on HTML and CSS'' in the resume, and dialog about the experience of web page development before, the system generates an interview question ``If a product needs a three-level classification selection, which component would you use and how to achieve it?'' is generated.

There are two main challenges in this task.
% We formulate the MIG task as a knowledge-based multi-turn dialog generation process, since the candidate resume and job description can be regarded as a form of knowledge.
One is that the knowledge-grounded interviews are extremely time-consuming and costly to collect. 
Without a sufficient amount of training data, the performance of such dialog generation models drops dramatically~\cite{zhao2020low}. 
The second challenge is to make the knowledge-grounded dialog relevant to the candidate resume, job description, and previous dialog utterances. 
This makes MIG a complex task involving text understanding, knowledge selection, and dialog generation.

%their use is considerably limited in a real-world setting as resume-grounded interview dialogs areAnd as the
% For example, the performance of an active learning model drops by 41.9\% when the size of dataset decreases from 10k to 1k \cite{yoo2019learning}.

% Hence, it is important to consider the MIG task in a low-resource setting, along with two further challenges:
% (1) the difficulty to understand the previous dialog utterance to generate a coherent and fluent interview conversation;
% (2) the challenge to select important, relevant points in the resume to conduct an examination with the job seeker.

In this paper, we propose EZInterviewer, a novel mock interview generator, with the aim of making interviews easier to prepare.
The key idea is to train EZInterviewer in a low-resource setting: the model is first pre-trained on large-scale ungrounded dialogs and resume data, and then fine-tuned on a very small set of resume-grounded interview dialogs.
% More concretely, we reduce the number of parameters in EZInterviewer that rely on resume-grounded interview dialogs by disentangling the knowledge selector and the dialog generator, so that most parameters in the model can be pre-trained with ungrounded dialogs and resume data that are much more readily available.% and no longer low-resource.
Specifically, the \textbf{knowledge selector} consists of a \textit{resume encoder} to encode the resume, and a \textit{key-value memory network} with mask self-attention mechanism, responsible for selecting relevant information in the resume to focus on to help generate the next interview utterance.
The \textbf{dialog generator} also has two components, a \textit{context encoder} which encodes the current dialog context, and a \textit{response decoder}, responsible for generating the next dialog utterance without knowledge from the resumes. 
This knowledge-insensitive dialog generator is coordinated with the knowledge selector by a \textbf{decoding manager} that dynamically determines which component is activated for utterance generation. 

It is noted that the number of parameters in the decoding manager can be small, therefore it only requires a small number of resume-grounded interview dialogs.
Extensive experiments on real-world interview dataset demonstrate the effectiveness of our model.

To summarize, our contributions are three-fold:

$\bullet$ We introduce a novel Mock Interview Generation task, which is a pilot study of intelligent online recruitment with potential commercial values.

$\bullet$ To address the low-resource challenge, we propose to reduce the number of parameters that rely on interview dialogs by disentangling knowledge selector and dialog generator so that the majority of parameters can be trained with large-scale ungrounded dialog and resume data.

$\bullet$ We propose a novel model to jointly process dialog contexts, candidate resumes, and job descriptions and generate highly relevant, knowledge-aware interview dialogs.

\section{Related Work}
\label{sec:related}

Multi-turn response generation aims to generate a response that is natural and relevant to the entire context, based on utterances in its previous turns.
\cite{zhang2019dialogpt} concatenated multiple utterances into one sentence and utilized RNN encoder or Transformer to encode the long sequence, simplifying multi-turn dialog into a single-turn dialog.
To better model the relationship between multi-turn utterances, \cite{gao2020learning,chen2020reasoning} introduced interaction between utterances after encoding each utterance.

% \noindent \textbf{Knowledge Selection in Dialog.}
As human conversations are almost always grounded with external knowledge, the absence of knowledge grounding has become one of the major gaps between current open-domain dialog systems and real human conversations~\cite{fu2019multiple,zhang2021rise,niu2021open}.
A series of work~\cite{li2019incremental,tian2020response} focused on generating a response based on the interaction between context and unstructured document knowledge, while a few others~\cite{xu2020knowledge,liu2020towards} introduced knowledge graphs into conversations.
These models, however, usually under-perform in a low-resource setting.

To address the low resource problem, \cite{li2020cross} proposed to enhance the context-dependent cross-lingual mapping upon the pre-trained monolingual BERT representations.
\cite{tae2020meta} extended the meta-learning algorithm, which utilized knowledge learned from high-resource domains to boost the performance of low-resource unsupervised neural machine translation.
Different from the above methods, \cite{zhao2020low} proposed a disentangled response decoder in order to isolate parameters that depend on knowledge-grounded dialogs from the entire generation model.
Our model takes a step further, taking into account the changes in attention on knowledge in multi-turn dialog scenarios.

\section{Model}
\subsection{Problem Formulation}
\label{sec:formulation}

For an input multi-turn dialog context $U=\{u_1, u_2, \dots, u_m\}$ between a job candidate and an interviewer, where $u_i$ represents the $i$-th utterance, we assume there is a ground truth textual interview question $Y=\{y_1, y_2, \dots, y_n\}$.
$m$ is the utterance number in the dialog context and $n$ is the total number of words in question $Y$.
In the $i$-th utterance, $u_i=\{x^i_1, x^i_2, \dots, x^i_{T_u^i}\}$.
Meanwhile, there is a candidate resume $R=\{(k_1, v_1), (k_2, v_2), \dots, (k_{T_r}, v_{T_r}) \}$ corresponding to the candidate in the interview, which has $T_r$ key-value pairs, and each of which represents an attribute in the resume.
For the job-resume matching pre-training task, there is an external job description $J=\{j_1, j_2, \dots, j_{T_j}\}$, which has $T_j$ words.
The goal is to generate an interview question $Y^{'}$ that is not only coherent with the dialog context $U$ but also pertinent to the job candidate's resume $R$.

\begin{figure*}
    \centering
    \includegraphics[scale=0.6]{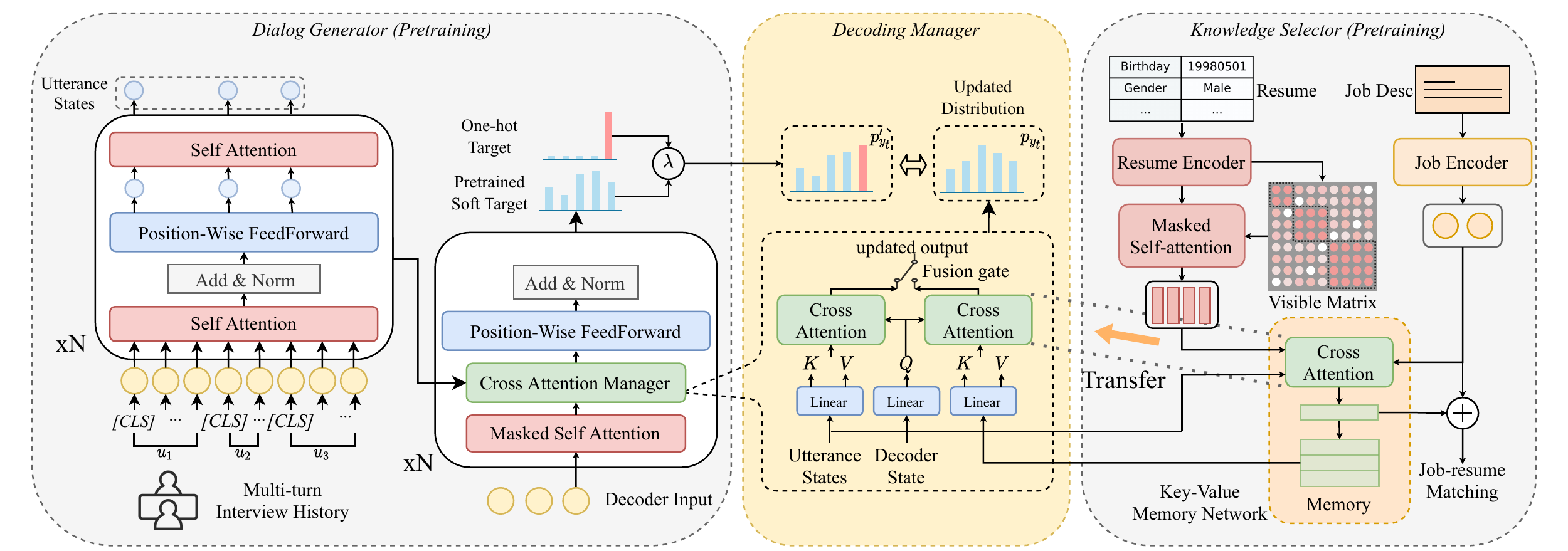}
    \caption{
        Overview of EZInterviewer, which consists of three parts: (1) \textit{Knowledge Selector} selects salient knowledge information from the candidate resume; (2) \textit{Dialog Generator} predicts the next word without knowledge of resumes; (3) \textit{Decoding Manager} coordinates the output from knowledge selector and dialog generator to produce the interview question.
    }
    \label{fig:overview}
\end{figure*}

\subsection{System Overview}
In this section, we propose our Low-resource Mock Interview Generator (EZInterviewer) model, which is divided into three parts as shown in Figure~\ref{fig:overview}:

% $\bullet$ \textit{Feature Encoder} is composed of a context encoder, which encodes the multi-turn dialog context, and a resume encoder, which encodes the candidate resume.
$\bullet$ \textit{Dialog Generator} predicts the next word of a response based on the prior sub-sequence.
In our model, we pre-train it by large-scale ungrounded dialogs.

$\bullet$ \textit{Knowledge Selector} selects salient knowledge information from the candidate resume for interview question generation.
In our model, we augment the ability of the knowledge selector by employing it to perform job-resume matching.

%$\bullet$ \textit{Knowledge Memory} memorizes the development of knowledge preference as the interview continues.

$\bullet$ \textit{Decoding Manager} coordinates the output from knowledge selector and dialog generator to predict the interview question.

It is important to note that to train an EZInterviewer model, two pre-train techniques are employed.
Firstly, we pre-train the knowledge selector in a job-matching task.
This is because while it is hard to attend to appropriate content in a resume just on its own, the salient information in a resume can be identified in a job-resume matching task~\cite{yan2019interview, le2019towards}.
Secondly, the context encoder and response decoder of the dialog generator are pre-trained with a large scale of ungrounded dialogs, so as to predict the next word of response based on the prior sub-sequence.
Finally, the decoding manager, which relies on a few parameters, coordinates the two components to generate knowledge grounded interview utterance.

% \subsection{Feature Encoder}
% \textbf{Context Encoder.}

% \textbf{Resume Encoder.}
% The information in a resume can be seen as structured knowledge data in our task.

\subsection{Dialog Generator}
\label{sec:dialog_generator}
\textbf{Context Encoder.}
Instead of processing the dialog context as a flat sequence, we employ a hierarchical encoder~\cite{chen2022target} to capture intra- and inter-utterance relations, which is composed of a local sentence encoder and a global context encoder.
For the sentence encoder, to model the semantic meaning of the dialog context, we learn the representation of each utterance $u_i$ by a self-attention mechanism (SAM) initialized by BERT \cite{Devlin2019BERTPO}:
\begin{align}
h^i_j = \text{SAM}_\text{u}(e(x^i_j), h^i_{*}).
\end{align}
We extract the state at ``[cls]'' position to denote the utterance state, abbreviated as $h^i$.
Apart from the local information exchange in each utterance, we let information flow across multi-turn context:
\begin{align}
h^c_t = \text{SAM}_\text{c}(h^t, h^c_{*}),
\end{align}
where $h^c_t$ denotes the hidden state of the $t$-th utterance in $\text{SAM}_\text{c}$.

\textbf{Response Decoder.}
Response decoder is responsible for understanding the previous dialog context and generates the response without the knowledge of resume information~\cite{li2022keywords}.
Our decoder also follows the style of Transformer.

Concretely, we first apply the self-attention on the masked decoder input, obtaining $d_t$.
Based on $d_t$ we compute the cross-attention scores over previous utterances:
\begin{align}
\alpha^c_t=\text{ReLU}([d_t W_d(h^{c}_{i} W_h)^T]).
\end{align}
The attention weights $\alpha^c_t$ is then used to obtain the context vectors as $c_{t}=\sum_{i=1}^{m} \alpha^c_t h^{c}_{i}$.
The context vectors $c_t$, treated as salient contents of various sources, are concatenated with the decoder hidden state $d_t$ to produce the distribution over the target vocabulary:
\begin{align}
    P_v^w=\text{Softmax}\left(W_{o}\left[d_{t} ; c_t\right]\right).
    \label{equ:p}
\end{align}

\textbf{Pre-training process.}
While interview dialogs are hard to come by, online conversation is abundant on the internet, and can be easily collected.
Hence, we pre-train the dialog generator on ungrounded conversations.
Concretely, during pre-training process, we employ the context encoder to first encode the multi-turn previous dialog context.
Then, at the $t$-th decoding step, we use the response decoder to predict the $t$-th word in the response.
We set the loss as the negative log likelihood of the target word $y_t$:
\begin{equation}\label{loss-g}
    Loss_g = - \frac{1}{n} \textstyle \sum^{n}_{t=1} \log P^w_{v}(y_t).
\end{equation}

\subsection{Knowledge Selector}
\label{sec:knowledge selector}
\textbf{Resume Encoder.}
As shown in Figure~\ref{fig:overview}, a resume contains several key-value pairs $(k_i, v_i)$.
Most of key and value fields include a single word or a phrase such as ``skills'' or ``gender'', and we can obtain the feature representation through an embedding matrix.
Concretely, for each key or value field with a single word or a phrase, we establish a corresponding resume embedding matrix $e_r^i$ that is different from the previous one.
Then we use the resume embedding matrix to map each field word $k_i$ or $v_i$ into to a high-dimensional vector space, denoted as $e_r^i(k_i)$ or $e_r^i(v_i)$.
For fields with more than one word such as ``work experience'' or ``I used to...'', we denote them as $v_i=(v_i^1,...v_i^{l_i})$, where $l_i$ denotes the word number of the current field.
We first process them through the previous word embedding matrix $e$, then there is an $\rm{\text{SAM}_R}$, similar with $\rm{\text{SAM}_u}$ in Section~\ref{sec:dialog_generator}, to model the temporal interactions between words:
\begin{align}
h^{r_i}_t = \text{SAM}_\text{R}(e(v_i^j), h^{r_i}_{t-1}).
\end{align}
We use the last hidden state of the $\text{SAM}_\text{R}$, \ie $h^{r_i}_{l_i}$ to denote the overall representation for field $v_i$.
% Generally, we represent the value $v_i$ as $h^v_i$.
% Similarity, we concatenate the last hidden state of the forward and backward to obtain the overall vector.
% We denote $h^v_i=e_i(k_i)$ as the feature representation of $v_i$, which is obtained from an embedding matrix or a text encoder.

For brevity, in the following sections, we use $h^k_i$ and $h^v_i$ to denote the encoded key-value pair $(k_i, v_i)$ in the resume.

% \textbf{Key-Value Memory Network.} As key-value memory network (KVMN) is shown effective in structured data utilization ~\cite{madotto2018mem2seq}, we employ KVMN to extract resume information, which includes two steps: key matching and value combination.

\textbf{Masked Self-attention.}
Traditional self-attention can be used to update representation of each resume item due to its flexibility in relating two elements in a distance-agnostic manner \cite{li2020vmsmo}.
However, as shown in \cite{liu2020k}, too much knowledge incorporation may divert the representation from its correct meaning, which is called knowledge noise (KN) issue.
In our scenario, the information in the resume is divided into several parts, \ie basic personal information, work experiences and extended work, each of which contains variable number of items.
The items within each part are closely connected, while different parts can be considered as different domains, and the interaction may introduce a certain amount of noise.
% To overcome this problem, we introduce a visible matrix, in which items belonging to the same part, \ie individual information, work experiences and expected position, are visible to each other， while the visibility degree between items from different parts is determined by the cosine similarity of semantic representations, \ie $C_{i,j}=\text{cos\_sim}(h^v_i, h^v_j)$.
To overcome this problem, we introduce a visible matrix, in which items belonging to the same part are visible to each other, while the visibility degree between items is determined by the cosine similarity of semantic representations, \ie $C_{i,j}=\text{cos\_sim}(h^v_i, h^v_j)$.
Then, the scaled dot-product masked self-attention is defined as:
\begin{align}
\alpha_{i, j} &=\frac{\exp \left((h^k_i W_q) C_{i,j} (h^k_j W_k)^T\right)}{\sum\nolimits_{n=1}^{T_{r}} \exp \left((h^k_i W_q) C_{i,n} (h^k_j W_k)^T\right)}, \\
\hat{h}^v_{i} &=\sum\nolimits_{j=1}^{T_{r}} \frac{\alpha_{i, j} h^v_j}{\sqrt d},
\end{align}
where $d$ stands for hidden dimension and $C$ is the visible matrix.
$\hat{h}^v_{i}$ is then utilized as the updated resume value representation.

\textbf{Key-Value Memory Network.}
The goal of key matching is to calculate the relevance between each attribute of the resume and the previous dialog context.
Given dialog context $h^i$, for the $j$-th attribute pair $(k_j, v_j)$, we calculate the probability of $h^i$ over $k_j$, \ie $P(k_j|h^i)$, as the \textit{matching score} $\beta_{i,j}$.
To this end, we exploit the context representation $h^i$ to calculate the matching score:
\begin{align}
\beta_{i,j} &=\frac{\exp \left(h^i W_a h^k_j\right)}{\sum\nolimits_{n=1}^{T_r} \exp \left(h^{i} W_a  h^k_{n}\right)}.
\label{equ:attn}
\end{align}
Since context representation $h^i$ and resume key representation $h^k_j$ are not in the same semantic space, we use a trainable key matching parameter $W_a$ to transform these representations into a same space.

% \textit{Value Combination.}
As the relevance between context $h^i$ and each pair in the resume table $(k_j, v_j)$, the matching score $\beta_{i,j}$ can help to capture the most relevant pair for generating a correct question.
Therefore, as shown in Equation~\ref{equ:value}, the knowledge selector reads the information $M_i$ from KVMN via summing over the stored values, and guides the follow-up response generation, so we have:
\begin{align}
\label{equ:value}
M_{i} &=\sum\nolimits_{j=1}^{T_{r}} \beta_{i, j} \hat{h}^v_j,
\end{align}
where $\hat{h}^v_j$ is the representation of value $v_j$, and $\beta_{i,j}$ is the matching score between dialog context $h^i$ and key $k_j$.

\textbf{Pre-training Process.}
In practice, the resume knowledge contains a variety of professional and advanced scientific concepts such as ``Web front-end'', ``HTML'', and ``CSS''.
These technical terms are difficult to understand for people not familiar with the specific domain, not to mention for the model that is not able to access a large-scale resume-grounded dialog dataset.
Hence, it would be difficult for the knowledge selector to understand the resume content and previous context about the resume, so as to select the next resume pair to focus on.

On the other hand, we notice that in job-resume matching task, it is crucial to capture the decisive information in the resume to perform a good matching.
For example, recruiters may tend to hire the candidate with particular experiences among several candidates with similar backgrounds~\cite{yan2019interview}.
Intuitively, the key-value pair that is important for job-resume matching is also the key factor to consider in a job interview.
Hence, if we can let the model learn the salient information in the resume by performing the job-resume matching task on large-scale job-resume data, then it would also bring benefits for selecting salient information in interview question generation.

Concretely, we use the job description to attend to the resume to perform a job-resume matching task, as a pre-training process for knowledge selector module.
As shown in Figure~\ref{fig:overview}, the \textit{Job Encoder} encodes the job description by a $\text{SAM}_\text{jd}$:
\begin{align}
h^{jd}_i = \text{SAM}_\text{jd}(e(j_i), h^{jd}_{i-1}),
\end{align}
where $j_i$ denotes the $i$-th word in the job description, and $e(j_i)$ is mapped by the previous embedding matrix $e$.
We use the final hidden state of the $\text{SAM}_\text{jd}$, \ie $h^{jd}_{T_j}$ as the overall representation for the description, abbreviated as $h^{jd}$.
$h^{jd}$ plays a similar part as the context representation $h^i$, which first attends to the keys in the resume,
% \begin{align}
% \alpha_{j} &=\frac{\exp \left(h^{jd} W_a h^k_j\right)}{\sum\nolimits_{n=1}^{T_r} \exp \left(h^{jd} W_a  h^k_{n}\right)}.
% \end{align}
and then is used to ``weightedly'' read the values in the resume.
We use $m_{jd}$ to denote the weighted read result.
% \begin{align}
% m_{jd} &=\sum\nolimits_{j=1}^{T_{r}} \alpha_{j} h^v_j.
% \end{align}

In the training process, we first pre-train the knowledge selector by job-resume matching task, which can be formulated as a classification problem~\cite{qin2018enhancing}.
% We calculate the job-resume matching score based on the previously-obtained job description representation $h^{jd}$ and selected resume knowledge $m_{jd}$.
The objective is to maximize the scores of positive samples while minimizing that of the negative samples.
Specifically, we concatenate $h^{jd}$ and $m_{jd}$ since vector concatenation for matching is known to be effective~\cite{severyn2015learning}.
Then the concatenated vector is fed to a multi-layer, fully-connected, feed-forward neural network, and the job-resume matching score $s_{jr}$ is obtained as:
\begin{align}
s_{jr} = \sigma\left( F_s([h^{jd};m_{jd}]]) \right),
\end{align}
where $[;]$ denotes concatenation operation, and the outputs are the probabilities of successfully matching.
We use the job-resume pairs in interviews as positive samples, and then use the job-resume pairs without interviews as negative instances.
% We apply the cross-entropy loss for training the pre-training job-resume networks.

After pre-training, the job description is replaced by the context representations, while the key matching and value combination processes remain the same.
% In this way, the knowledge selector learns how to calculate the relationship between keys and the job description (dialog context).
We use a knowledge memory $M$ to store the selection result, where each slot stores the value combination result $M_i$ in Equation~\ref{equ:value}.

% \subsection{Knowledge Memory}

\subsection{Decoding Manager}
The decoding manager is supposed to generate the proper word based on the knowledge memory and the response decoder.
Our idea is inspired by an observation on the nature of interview dialogs: despite the fact that a dialog is based on the resume, words and utterances in the dialog are not always related to resume.
Therefore, we postulate that formation of a response can be decomposed into two uncorrelated actions:
(1) selecting a word according to the context to make the dialog coherent (corresponding to the dialog generator);
(2) selecting a word according to the extra knowledge memory to ground the dialog (corresponding to the knowledge selector).
The two actions can be independently performed, which becomes the key reason why the large resume-job matching and ungrounded dialog datasets, although seemingly unrelated to interview dialogs, can be very useful in an MIG task.

Note that in Section \S\ref{sec:knowledge selector}, we store the selected knowledge $M_i$ in a knowledge memory $M$.
To select a word based on it, similar to the response decoder, we use $d_{t}$ to attend to each slot of knowledge memory, and we can obtain the knowledge context vector $g^k_t$ and the output decoder state $d^{ko}_t$.
% \begin{align}
%     \beta^{k'}_{t, i} &= z_{k}^\intercal \tanh \left( W^k_s K_i + W^k_d d_t \right), \\
%     \beta^k_{t, i} &= \exp \left( \beta^{k'}_{t, i} \right) /  \sum^{T_r}_{j=1} \exp \left(\beta^{k'}_{t, j} \right), \\
%     g^k_t &= \textstyle \sum_{i=1}^{T_r} \beta^k_{t, i} K_i.
% \end{align}

The response decoder and knowledge selector are controlled by the decoding manager with a ``fusion gate'' to decide how much information from each side should be focused on at each step of interview question prediction.
\begin{align}
    \gamma_f &=\sigma\left( F_m(d_t) \right),
\end{align}
where $d_t$ is the $t$-th decoder hidden state.
Then, the probability to predict word $y_t$ can be formulated as:
\begin{align}
    d^o_t &= \gamma_f d^{wo}_t + (1-\gamma_f) d^{ko}_t,\\
    P_{v} &= \text{softmax} \left(W_v d^{o}_t + b_v \right).
\end{align}

As for the optimization goal, generation models that use one-hot distribution optimization target always suffer from the over-confidence issue, which leads to poor generation diversity \cite{wang2021diversifying}.
% and the degradation problem is serious during finetune process of the pre-training model.
Hence, aside from the ground truth one-hot label $P$, we also propose a soft target label $P^w_v$ (see Equation \ref{equ:p}), which is borrowed from the pre-trained Dialog Generator in Section \ref{sec:dialog_generator}.
Forcing the decoding manager to simulate the pre-trained decoder can help it learn the context of the interview dialog.
We combine the one-hot label with the soft label by an editing gate $\lambda$, as shown in Figure~\ref{fig:overview}.
% , and the smooth target is calculated as:
Concretely, a smooth target distribution $P'$ is proposed to replace the hard target distribution $P$ as:
\begin{align}
    P' = \lambda P + (1-\lambda) P^w_v.
    \label{equ:soft}
\end{align}
where $\lambda \in [0,1]$ is an adaption factor, $P^w_v$ is obtained from Equation~\ref{equ:p}, and $P$ is the hard target as one-hot distribution which assigns a probability of 1 for the target word $y_t$ and 0 otherwise.

\section{Experimental Setup}
\label{sec:exp}
\subsection{Dataset}
\label{dataset}
In this paper, we conduct experiments on a real-world dataset provided by ``Boss Zhipin'' \footnote{\url{https://www.zhipin.com}}, the largest online recruiting platform in China.
To protect the privacy of candidates, user records are anonymized with all personal identity information removed.
% With all the data obtained, we first take multiple steps to cleanse data including removing incomplete resumes and interview conversations for a given period of time.
% The cleansed dataset includes 12,666 resumes, 8,032 job descriptions and 49,214 interview dialog utterances.
The dataset includes 12,666 resumes, 8,032 job descriptions, and 49,214 interview dialog utterances.
The statistics of the dataset is summarized in Table~\ref{tab:dataset}.
We then tokenize each sentence into words with the benchmark Chinese tokenizer toolkit ``JieBa''~\footnote{\url{https://github.com/fxsjy/jieba}}.

To pre-train the knowledge selector module, we use a job-resume matching dataset~\cite{yan2019interview}, again from ``Boss Zhipin''.
The training set and the validation set include 355,000 and 1,006 job-resume pairs, respectively.
To pre-train dialog generator, we choose Weibo dataset~\cite{chan2019modeling}, which includes a massive number of multi-turn conversations collected from ``Weibo''\footnote{\url{https://www.weibo.com}}.
The data includes 2,990,000 context-response pairs for training and 5,000 pairs for validation.
The details are also summarized in Table~\ref{tab:dataset}.
% On average, each context consists of 3 utterances.

\begin{table}[t]
    \centering
    \caption{Statistics of the datasets used in the experiments.}
    \small
    \begin{tabular}{@{}|l|r|@{}}
        \toprule
        Statistics & Values\\
        \midrule
        \multicolumn{2}{|l|}{Interview Dialog Dataset}\\
        \midrule
        Total number of resumes & 12,666\\
        Total number of dialog utterances & 49,214\\
        Avg turns \# per dialog context & 4.47\\
        Avg words \# per utterance & 13.18\\
        \midrule
        \multicolumn{2}{|l|}{Job-Resume Dataset}\\
        \midrule
        Key-value pairs \# per resume & 22\\
        Avg words \# per work experience in resume & 72.80\\
        Avg words \# per self description in resume & 51.13\\
        Avg words \# per job description & 74.26\\
        \midrule
        \multicolumn{2}{|l|}{Ungrounded Dialog Dataset}\\
        \midrule
        Total number of context-response pairs & 2,995,000\\
        Avg turns \# per dialog context & 4\\
        Avg words \# per utterance & 15.15\\
        \bottomrule
    \end{tabular}
    \label{tab:dataset}
\end{table}

\begin{table*}[t]
    \centering
    \small
    \caption{Comparing model performance on full dataset: automatic evaluation metrics.}
    \begin{tabular}{@{}l|cccc|ccc|cc|cc@{}}
      \toprule
      & BLEU-1 & BLEU-2 & BLEU-3 & BLEU-4 & Extrema & Average & Greedy & Dist-1 & Dist-2 & Entity F1 & Cor\\
      \midrule
      \multicolumn{3}{l}{\emph{Knowledge-insensitive dialog generation}}\\
      \midrule
    % Seq2Seq~\cite{Bahdanau2015NeuralMT} & 0.5256 & 0.3595 & 0.2485 & 0.2159 & 0.4840 & 0.7841 & 0.6805 & 0.0873 & 0.3582 & 0.3389 & 0.2580 \\
    % VAE~\cite{Zhao2017LearningDD} & 0.5346 & 0.3726 & 0.2620 & 0.2283 & 0.4837 & 0.7851 & 0.6674 & 0.0970 & 0.3702 & 0.3549 & 0.2870\\
    Transformer~\cite{vaswani2017attention} & 0.5339 & 0.3811 & 0.2836 & 0.2530 & 0.4859 & 0.7673 & 0.6803 & 0.0928 & 0.3157 & 0.3606 & 0.2711\\
    BERT~\cite{Devlin2019BERTPO} & 0.5671 & 0.3864 & 0.2735 & 0.2583 & 0.4861 & 0.7669 & 0.6792 & 0.0947 & 0.3558 & 0.3711 & 0.2894\\
    DialoGPT~\cite{zhang2019dialogpt} & 0.5722 & 0.4015 & 0.3004 & 0.2697 & 0.4858 & 0.7670 & 0.6814 & 0.1001 & 0.3620 & 0.3843 & 0.3002\\
    T5-CLAPS~\cite{Claps} & 0.5846 & 0.4126 & 0.3020 & 0.2783 & 0.4837 & 0.7851 & 0.6674 & 0.0970 & 0.3702 & 0.3549 & 0.2870\\
      \midrule
      \multicolumn{3}{l}{\emph{Knowledge-aware dialog generation}}\\
      \midrule
      TMN~\cite{dinan2018wizard} & 0.5437 & 0.3891 & 0.2963 & 0.2630 & 0.4841 & 0.7655 & 0.6811 & 0.0996 & 0.3299 & 0.3830 & 0.2652 \\
      ITDD~\cite{li2019incremental} & 0.5484 & 0.4009 & 0.2929 & 0.2656 & 0.4833 & 0.7650 & 0.6859 & 0.1055 & 0.3703 & 0.3661 & 0.2715 \\
      DiffKS~\cite{zheng2020difference} & 0.5617 & 0.3898 & 0.2776 & 0.2441 & 0.4826 & 0.7830 & 0.6752 & 0.0937 & 0.3612 & 0.3672 & 0.2750\\
      DRD~\cite{zhao2020low} & 0.5711 & 0.4001 & 0.2914 & 0.2548 & 0.4824 & 0.7813 & 0.6783 & 0.0867 & 0.3661 & 0.3825 & 0.2883 \\
      %\midrule
      %\emph{Table-based dialog generation}\\
      DDMN~\cite{wang2020dual} & 0.5693 & 0.4065 & 0.2968 & 0.2694 & 0.4831 & 0.7655 & 0.6811 & 0.0944 & 0.3640 & 0.3754 & 0.2869 \\
      Persona~\cite{fu2022there} & 0.5532 & 0.3829 & 0.2715 & 0.2377 & 0.4823 & 0.7822 & 0.6783 & 0.0911 & 0.3598 & 0.3833 & 0.2928 \\
      \midrule
      \textbf{EZInterviewer} & \textbf{0.6106} & \textbf{0.4320} & \textbf{0.3284} & \textbf{0.2917} & \textbf{0.4893} & \textbf{0.7884} & \textbf{0.6886} & \textbf{0.1071} & \textbf{0.3747} & \textbf{0.3927} & \textbf{0.3145}\\
      \quad No Pre-train & 0.5738 & 0.4029 & 0.2929 & 0.2599 & 0.4846 & 0.7833 & 0.6831 & 0.0981 & 0.3673 & 0.3819 & 0.3007 \\
      \quad w/o KM & 0.5795 & 0.4127 & 0.3069 & 0.2754 & 0.4847 & 0.7841 & 0.6762 & 0.0979 & 0.3685 & 0.3803 & 0.3010 \\
      \quad w/o KS & 0.5775 & 0.4122 & 0.3067 & 0.2746 & 0.4781 & 0.7668 & 0.6787 & 0.1003 & 0.3691 & 0.3848 & 0.2994 \\
      \quad w/o LS & 0.6007 & 0.4232 & 0.3176 & 0.2821 & 0.4869 & 0.7863 & 0.6832 & 0.0969 & 0.3664 & 0.3902 & 0.3127\\

      \bottomrule
    \end{tabular}
    \label{tab:baselines}
  \end{table*}

\subsection{Comparisons}
\label{comparison}

We compare our proposed model against traditional knowledge-insensitive dialog generation baselines, and knowledge-aware dialog generation baselines.
% traditional dialog generation baselines, document-based dialog generation baselines, and table-based dialog generation baselines.

 \noindent $\bullet$ \textbf{\textit{Knowledge-insensitive dialog generation baselines}}:

% \textbf{Seq2Seq}~\cite{Bahdanau2015NeuralMT}: the vanilla schema of the sequence to sequence model with attention mechanism.
\textbf{Transformer}~\cite{vaswani2017attention}: is based solely on attention mechanisms.
\textbf{BERT}~\cite{Devlin2019BERTPO}: initializes Transformer with BERT as the encoder.
\textbf{DialoGPT}~\cite{zhang2019dialogpt}: proposes a large, tunable neural conversational response generation model trained on more conversation-like exchanges.
\textbf{T5-CLAPS}~\cite{Claps}: generates samples for contrastive learning by adding small and large perturbations, respectively.

 \noindent $\bullet$ \textbf{\textit{Knowledge-aware dialog generation baselines}}:

\textbf{TMN}~\cite{dinan2018wizard}: is built upon a transformer architecture with an external memory hosting the knowledge.
\textbf{ITDD}~\cite{li2019incremental}: incrementally encodes multi-turn dialogs and knowledge and decodes responses with a deliberation technique.
\textbf{DiffKS}~\cite{zheng2020difference}: utilizes the differential information between selected knowledge in multi-turn conversation for knowledge selection.
\textbf{DRD}~\cite{zhao2020low}: tackles the low-resource challenge with pre-training techniques using ungrounded dialogs and documents.
\textbf{DDMN}~\cite{wang2020dual}: dynamically keeps track of dialog context for multi-turn interactions and incorporates KB knowledge into generation.
\textbf{Persona}~\cite{fu2022there}: introduces personal memory into knowledge selection to address the personalization issue.

\subsection{Implementation Details}

We implement our experiments in TensorFlow~\cite{Abadi2016TensorFlowAS} on an NVIDIA GTX 1080 Ti GPU.
For our model and all baselines, we follow the same setting as described below.
We truncate input dialog to 100 words with 20 words in each utterance, as we did not find significant improvement when increasing input length from 100 to 200 tokens.
The minimum decoding step is 10, and the maximum step is 20.
The word embedding dimension is set to 128 and the number of hidden units is 256.
% We initialize all of the parameters randomly using a Gaussian distribution.
Experiments are performed with a batch size of 256, and the vocabulary is comprised of the most frequent 50k words.
We use Adam optimizer~\cite{kingma2014adam} as our optimizing algorithm.
% We also apply gradient clipping~\cite{Pascanu2013OnTD} with a range of $[-2,2]$ during training.
% During the inference stage, the checkpoint with the smallest validation loss is chosen and and the beam-search size is set to 4 for all methods.
We selected the 5 best checkpoints based on performance on the validation set and report averaged results on the test set.
Note that for better performance, our model is built based on BERT, and the decoding process is the same as Transformer~\citep{vaswani2017attention}.
Finally, due to the limitation of time and memory, small settings are used in the pre-trained baselines.

% \begin{table*}[t]
%     \centering
%     \small
%     \caption{BLEU and Embedding scores comparison for EZInterviewer with different scale of data. }
%     \begin{tabular}{@{}lccccccccccc@{}}
%       \toprule
%       & BLEU-1 & BLEU-2 & BLEU-3 & BLEU-4 & Extrema & Average & Greedy & Dist-1 & Dist-2 & Entity F1 & Cor\\
%       \midrule
%       \textbf{Full data} & \textbf{0.6011} & \textbf{0.4264} & \textbf{0.3205} & \textbf{0.2883} & \textbf{0.4871} & \textbf{0.7872} & \textbf{0.6872} & \textbf{0.1069} & \textbf{0.3716} & 0.3906 \\
%     1/2 data & 0.5979 & 0.4253 & 0.3169 & 0.2832 & 0.4856 & 0.7862 & 0.6859 & 0.1052 & 0.3709 & \textbf{0.3912} \\
%     1/4 data & 0.5949 & 0.4215 & 0.3128 & 0.2804 & 0.4864 & 0.7855 & 0.6825 & 0.1047 & 0.3689 & 0.3898 \\
%     1/8 data & 0.5931 & 0.4195 & 0.3127 & 0.2795 & 0.4843 & 0.7859 & 0.6806 & 0.1054 & 0.3693 & 0.3860 \\
%     1/10 data & 0.5930 & 0.4195 & 0.3106 & 0.2779 & 0.4857 & 0.7851 & 0.6793 & 0.1051 & 0.3697 & 0.3864 \\
%       \bottomrule
%     \end{tabular}
%     \label{tab:different_scale}
%   \end{table*}

\subsection{Evaluation Metrics}
To evaluate the performance of  EZInterviewer against baselines, we adopt the following metrics widely used in existing studies.

\textbf{Overlap-based Metric.}
Following \cite{li2021style}, we utilize BLEU score \cite{papineni2002bleu} to measure n-grams overlaps between ground-truth and generated response.
% Following \cite{li2021style}, we utilize BLEU score \cite{papineni2002bleu}, widely used in machine translation and dialog system, to measure n-grams overlaps between ground-truth and generated text.
% BLEU-1, BLEU-2, BLEU-3, and BLEU-4 correspond to uni-gram, bi-gram, tri-gram, and 4-gram, respectively
% Specifically, following the conventional setting in previous work \cite{gu2018dialogwae}, we compute BLEU scores using smoothing techniques (smoothing 7).
In addition, we apply Correlation (Cor) to calculate the words overlap between generated question and job description, which measures how well the generated questions line up with the recruitment intention.

\textbf{Embedding Metrics.}
We compute the similarity between the bag-of-words (BOW) embeddings of generated results and reference to capture their semantic matching degrees~\cite{gu2018dialogwae}.
In particular we adopt three metrics: 1) \textit{Greedy}, \ie greedily matching words in two utterances based on cosine similarities;
2) \textit{Average}, cosine similarity between the averaged word embeddings in two utterances \cite{mitchell2008vector};
3) \textit{Extrema}, cosine similarity between the largest extreme values among the word embeddings in the two utterances \cite{forgues2014bootstrapping}.

\begin{table}[t]
    \centering
    \small
    \caption{Human evaluation results on: Readability (Read), Informativeness (Info), Meaningfulness (Mean), Usefulness (Use), Relevance (Rel), and Coherence (Coh).}
    % The paired student t-test between our model and DRD (the row with shaded background) provides further proof of the significance of these results.}
    \begin{tabular}{@{}l|cc|cccc@{}}
      \toprule
      \multirow{2}{*}{Model} & \multicolumn{2}{c|}{Dialog-level} & \multicolumn{4}{c}{Interview-level}\\
  \cline{2-7}
      & Read & Info & Mean & Use & Rel & Coh\\
      \midrule
      DiffKS & 1.79 & 2.01 & 1.87 & 2.03 & 1.99 & 2.10\\
      DDMN & 1.97 & 1.83 & 1.63 & 2.12 & 2.14 & 1.91\\
      \cbkgrnd DRD &\cbkgrnd 2.05  &\cbkgrnd 2.11 & \cbkgrnd 2.09 & \cbkgrnd 2.08 & \cbkgrnd 2.17 & \cbkgrnd 2.02\\
      EZInterviewer & \textbf{2.42}\dubbelop &\textbf{2.51}\dubbelop  & \textbf{2.39}\dubbelop & \textbf{2.46}\dubbelop & \textbf{2.57}\dubbelop & \textbf{2.38}\dubbelop\\
      \bottomrule
    \end{tabular}
    \label{tab:human_evaluation}
  \end{table}

\textbf{Distinctness.}
The distinctness score~\cite{Li2016ADO} measures word-level diversity by calculating the ratio of distinct uni-gram and bi-grams in generated responses.

\textbf{Entity F1.}
Entity F1 is computed by micro-averaging precision and recall over knowledge-based entities in the entire set of system responses, and evaluates the ability of a model to generate relevant entities to achieve specific tasks from the provided knowledge base~\cite{wang2020dual}.
The entities we use are extracted from an entity vocabulary provided by ``Boss Zhipin''.
% It is worth noting that entity F1 indicates the task-completion ability of a model, since knowledge-based entities are key in a dialog generating task.

\textbf{Human Evaluation Metrics.}
We further employ human evaluations aside from automatic evaluations.
Three well-educated annotators from different majors are hired to evaluate the quality of generated responses, where the evaluation is conducted in a double-blind fashion.
In total 100 randomly sampled responses generated by each model are rated by each annotator on both dialog level and interview level.
We adopt the \textit{Readability} (is the response grammatically correct?) and \textit{Informativeness} (does the response include informative words?) to judge the quality of the generated responses on the dialog level.
On the interview level, we adopt
\textit{Meaningfulness} (is the generated question meaningful?), \textit{Usefulness} (is the question worth the job candidate preparing in advance?), \textit{Relevance} (is the question relevant to the resume?) and \textit{Coherence} (is the generated text coherent with the context?) to assess the overall performance of a model and the quality of user experience.
Each metric is given a score between 1 and 3 (1 = bad, 2 = average, 3 = good).

\section{Experimental Result}
\label{sec:analysis}

\subsection{Overall Performance}
\textbf{Automatic evaluation.}
The comparison between EZInterviewer and state-of-the-art generative baselines is listed in Table~\ref{tab:baselines}.

We take note that the knowledge-aware dialog generation models outperform traditional dialog models, suggesting that utilizing external knowledge introduces advantages in generating relevant response.
We also notice the pre-train based model DRD outperforms other baselines, showing that initializing parameters by pre-training on large-scale data can lead to a substantial improvement in performance.
It is worth noting some models achieve better Entity F1 but a lower BLEU score; this suggests that those models tend to copy necessary entity words from the knowledge but are not able to use them properly.

EZInterviewer outperforms baselines on all automatic metrics.
Firstly, our model improves BLEU-1 by 6.92\% over DRD.
On the Distinctness metric Dist-1, our model outperforms DialoGPT by 6.99\%, suggesting that the generated interview questions are diversified and personalized with different candidates' resumes.
Moreover our model attains a good score of 0.3927 on entity F1, which evaluates the degree to which the generated question is grounded on the knowledge base.
Finally, Cor score of 0.3145 suggests the questions generated by EZInterviewer is in line with the job description, hence reflect the intention of the recruiters.
Overall the metrics demonstrate that our model successfully learns an interviewer's points of interest in a resume, and incorporates this knowledge into interview questions properly.
% This demonstrates that our model generates more appropriate responses by understanding the dialog context history, and selects resume knowledge better by pretraining on job-resume matching.

\textbf{Human evaluation.}
The results of human evaluations on all models are listed in Table~\ref{tab:human_evaluation}.
EZInterviewer is the top performer on all the metrics.
Specifically, our model outperforms DiffKS by 35.20\% on Readability, suggesting that EZInterviewer manages to reduce the grammatical errors and improve the readability of the generated response.
As for the Informativeness metric, our model scores 0.68 higher than DDMN.
This indicates that EZInterviewer captures salient information in the resume.
On the interview level, EZInterviewer's Usefulness score is 18.27\% better than DRD, demonstrating its capabilities to help job seekers to pick the right questions to prepare.
On Relevance metric, our model outperforms all baselines by a considerable margin, suggesting that the generated questions are closely related to the interview process.
Our model also performs better than other baselines in Meaningfulness and Coherence metrics, suggesting the overall higher quality of our model.

The above results demonstrate the competence of EZInterviewer in producing meaningful and useful interview questions whilst keeping the interview dialog flowing smoothly, just like a human recruiter.
Note that the average kappa statistics of human evaluation are 0.51 and 0.48 on dialog level and interview level, respectively, which indicates moderate agreement between annotators.
To prove the significance of these results, we also conduct the two-tailed paired student t-test between our model and DRD (row with shaded background).
The statistical significance of observed differences is denoted using \dubbelop (or \dubbelneer) for strong (or weak) significance for $\alpha = 0.01$.
Moreover, we obtain an average p-value of $5 \times 10^{-6}$ and $3 \times 10^{-4}$ for both levels, respectively.

\begin{figure}
    \centering
    \includegraphics[scale=0.5]{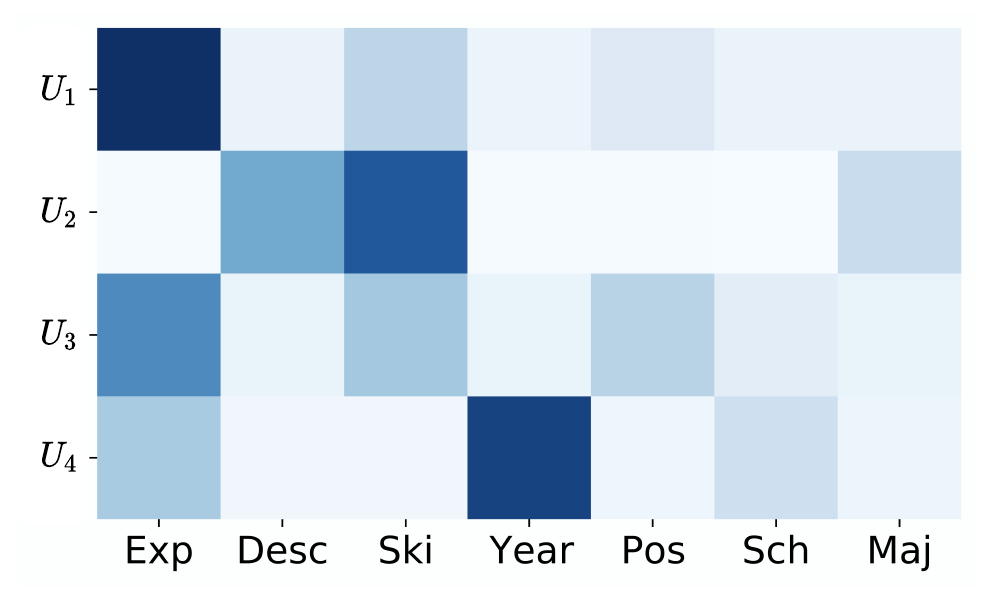}
    \caption{
        Visualization of key matching between dialog context and selected resume keys, \ie work experiment (Exp), self description (Desc), skills (Ski), work years (Year), expected position (Pos), school (Sch), and major (Maj). $U_i$ denotes the $i$-th utterance.
    }
    \label{fig:attn_matrix}
\end{figure}

\subsection{Ablation Study}
We conduct an ablation study to assess the contribution of individual components in the model. The results are shown in Table~\ref{tab:baselines}.

To verify the effectiveness of knowledge memory, we omit the knowledge selection of dialog context history and directly use the last utterance representation to select knowledge.
The results (see row w/o KM) confirm that employing each turn of historical dialog to select knowledge and saving it in memory contribute to generating better responses.
To confirm whether selecting knowledge helps with the response generation process, we remove it from the model, then simply add the representation of each utterance with all resume values, and store it into the memory.
This results in a drop of 5.42\% in BLEU-1 (see row w/o KS),  suggesting that selecting resume knowledge is beneficial in response generation.

\begin{table*}[t]
    \centering
    \small
    \caption{Translated interview questions generated by baselines and EZInterviewer: an example.
    \hlc[pink]{\quad\quad\quad\quad} denotes information extracted and words generated by knowledge selector, whereas \hlc[cyan!50]{\quad\quad\quad\quad} denotes words generated from dialog generator.}
    \begin{tabular}{@{}|l|l|l|@{}}
    %   \toprule
      \cline{1-3}
      \multicolumn{2}{|c|}{Resume} & \multicolumn{1}{c|}{Interview}\\
      \cline{1-3}
      Gender & Male & \multirow{4}{9.5cm}{Job Description:\\The main content of this work includes design and development based on the React front-end framework. It requires the ability to efficiently complete front-end development work and serve customers well.}\\
      \cline{1-2}
      Age & 28 & \\
      \cline{1-2}
      Education & Undergraduate & \\
      \cline{1-2}
      Major & Computer Science & \\
      \cline{1-3}
      Work Years & 10 & \multirow{5}{9.5cm}{Context:\\
      U1: Have you been engaged in front-end development work before?\\
      U2: Yes, I am good at Vue, Node.js and some other skills.\\
      U3: Okay, so do you have any React related experience?\\
      U4: Yes, I have more than 10 years of work experience.}\\
      \cline{1-2}
      Expected Position & Front-end Engineer &\\
      \cline{1-2}
      Low Salary & 5 &\\
      \cline{1-2}
      High Salary & 6 &\\
      \cline{1-2}
      Skills & Vue, Node.js, Java &\\
      \cline{1-3}
      \hlc[pink]{Experience} & \multicolumn{1}{p{5cm}|}{I was engaged in \hlc[pink]{front-end design} and was responsible for the project development based on the \hlc[pink]{React front-end framework} and participated in the system architecture process.} & \multirow{4}{9.5cm}{Ground Truth: So can you introduce a React related project you have done?\\
      DDMN: What other front-end frameworks would you use?\\
      DRD: Hello, can you tell us about your previous work?\\
      EZInterviewer: \hlc[cyan!50]{Well, can you introduce the} \hlc[pink]{experience} \hlc[cyan!50]{based on} \hlc[pink]{React framework?}}\\
      \bottomrule
    \end{tabular}
    \label{tab:case_study}
  \end{table*}

 \begin{figure*}
    \centering
    \includegraphics[scale=0.5]{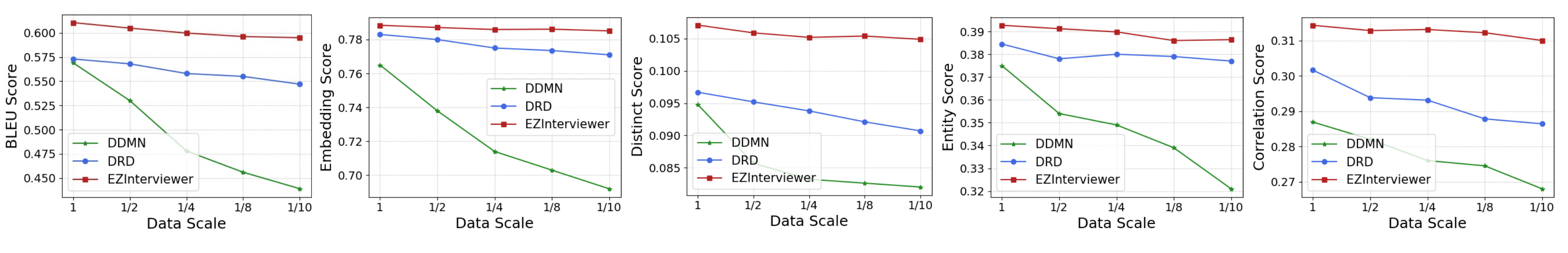}
    \caption{
        Automatic evaluation metrics of DDMN, DRD and EZInterviewer on training data of different scales.
    }
    \label{fig:bleu_scale}
\end{figure*}

\subsection{Analysis of Knowledge Selector}

In Section \S~\ref{sec:knowledge selector}, we introduce the selecting mechanism of knowledge selector, where the final attention (matching) score is obtained in Equation~\ref{equ:attn}.
To study what specific information is attended by the knowledge selector, and whether the selected information is suitable for the next interview question, we conduct a case study to visualize the matching score produced by the knowledge selector, as shown in Table~\ref{tab:case_study} and Figure~\ref{fig:attn_matrix}.
The first utterance in the history is ``Have you been engaged in front-end development work before?'', and the knowledge selector learns that this utterance focuses on the work experience in the resume.
Accordingly, the fourth utterance ``I have more than 10 years of work experience.'' pays more attention to work years and work experience than other items in the resume.
This demonstrates that the knowledge selector learns which item in the resume to focus on when generating each utterance.
Hence, when we want to ask the candidate to ``introduce a React related project'', the knowledge selector focuses on the work experience in the resume and generates the mock interview question.

\subsection{Impact of Training Data Scales}
To understand how our model and baseline models perform in a low-resource scenario, we first evaluate them on the full training dataset, then on smaller portions of the training dataset.
Figure~\ref{fig:bleu_scale} presents the performance of the models, DDMN, DRD, and EZInterviewer, on the full, 1/2, 1/4, 1/8 and 1/10 of the training dataset (data scale), respectively.
It is observed that as the size of training dataset reduces, DDMN suffers a massive drop across all metrics, whereas the scores of pre-training based models, \ie DRD and EZInterviewer, stay relatively stable.
This demonstrates pre-training as an effective strategy to tackle the low-resource challenge.
Moreover, our model outperforms DRD on all data scales, demonstrating the superiority of our model.
Figure~\ref{fig:bleu_scale} shows EZInterviewer eventually achieves the best performance on all metrics and outperforms (albeit slightly), with only \textbf{1/10} training data against all state-of-the-art baselines trained with the full training dataset.

\subsection{Case Study}
Table~\ref{tab:case_study} presents a translated example of EZInterviewer and baseline models.
We observe that the question from EZInterviewer not only catches the context, but also expands the conversation with proper knowledge.
This is highlighted in color codes: pink-colored words, \ie ``experience'' and ``React framework'', are what knowledge selector extracts from resume knowledge, whereas blue-colored words, \ie ``Well, can you introduce the...'' and ``based on'', which closely connect to the context, are generated by dialog generator.
% In contrast, the questions from the baselines such as ``tell us about your previous work'' and ``what other frameworks would you use'' respond to the dialog but fail to make connection with the resume knowledge.
In contrast, the questions from the baselines respond to the dialog but fail to make connection with the resume knowledge.

\section{Conclusion}
\label{sec:conclusion}
In this paper, we conduct a pilot study for the novel application of intelligent online recruitment, namely EZInterviewer, which aims to serve as mock interviewers for job-seekers.
The mock interview is generated with thorough understanding of the candidate's resume, the job requirements, the previous utterances in the context, as well as the selected knowledge for grounded interviews.
To address the low-resource challenge, EZInterviewer is trained on a very small set of interview dialogs.
The key idea is to reduce the number of parameters that rely on interview dialogs by disentangling the knowledge selector and dialog generator so that most parameters can be trained with ungrounded dialogs as well as the resume data that are not low-resource.
We conduct extensive experiments to demonstrate the effectiveness of the proposed solution EZInterviewer. Our model achieves the best results using full training data as well as small subsets of the training data in terms of various metrics such as BLEU, embedding based similarity and diversity, as well as human judgments.
In particular, the human evaluation indicates that our solution EZInterviewer can provide satisfactory mock interviews to help the job-seekers prepare the real interview, making the interview preparation process easier.

%  As a step towards making mock interview a reality, we introduce Mock Interview Generation (MIG) task in this paper, the first kind of dialog generation task in an interview setting.
%  We then propose a Low-resource Mock Interview Generator (EZInterviewer) model to further address the main challenge that all MIG tasks face: lack of sufficient resume-grounded interview dialogs.
%  EZInterviewer tackles this low-resource challenge by taking advantage of readily available ungrounded dialogs and resume corpus: its encoders and decoders are decoupled, pre-trained with this ungrounded data, and then re-integrated with the help of a special decoding manager.
%  This novel approach greatly reduces number of parameters that are reliant on resume-grounded interview dialogs, enabling the model to attain a same level of performance, on a considerably smaller (1/10) training dataset, to state-of-the-art baselines trained on the full dataset.
%  It is further shown that when trained with the same full set of interview dialogs, EZInterviewer model outperforms all state-of-the-art baselines by a notable margin across a range of human and automatic evaluation metrics.
%  We are now working to bring EZInterviewer online in the near future.

\section*{Acknowledgments}
We would like to thank the anonymous reviewers for their constructive comments. 
% We would also like to thank Anna Hennig in Inception Institute of Artificial Intelligence for her help on this paper. 
This work was supported by National Natural Science Foundation
of China (NSFC Grant No. 62122089).
Rui Yan is supported by Beijing Academy of Artificial Intelligence (BAAI).
% Rui Yan is partially supported as a Young Fellow of Beijing Institute of Artificial Intelligence (BAAI).

\end{CJK*}

\bibliographystyle{ACM-Reference-Format}
\balance
\bibliography{wsdm2023}

\end{document}